\pdfoutput=1

\documentclass[11pt]{article}

\usepackage[preprint]{acl}

\usepackage{times}
\usepackage{latexsym}
\usepackage{subcaption}

\usepackage[T1]{fontenc}

\usepackage[utf8]{inputenc}

\usepackage{microtype}

\usepackage{inconsolata}

\usepackage{graphicx}

\title{BPQA Dataset: Evaluating How Well Language Models \\Leverage Blood Pressures to Answer Biomedical Questions}

\author{Chi Hang \\
  NYU Center for Data Science / ch3946@nyu.edu \\
  Affiliation / Address line 2 \\
  Affiliation / Address line 3 \\
  \texttt{email@domain} \\\And
  Second Author \\
  Affiliation / Address line 1 \\
  Affiliation / Address line 2 \\
  Affiliation / Address line 3 \\
  \texttt{email@domain} \\}

\author{
  \textbf{Chi Hang\textsuperscript{1,3}},
  \textbf{Ruiqi Deng\textsuperscript{1,3}},
  \textbf{Lavender Yao Jiang\textsuperscript{1,3}},
  \textbf{Zihao Yang\textsuperscript{1,3}},
\\
  \textbf{Anton Alyakin\textsuperscript{3,4}},
  \textbf{Daniel Alber\textsuperscript{2,3}},
  \textbf{Eric Karl Oermann\textsuperscript{2,3}}
\\
\\
  \textsuperscript{1}NYU Center for Data Science,
  \textsuperscript{2}NYU Grossman School of Medicine,
\\
  \textsuperscript{3}NYU Langone Health,
  \textsuperscript{4}Washington University, Saint Louis
\\
  \small{\texttt{\{ch3946, rd2755, lyj2002, gavin.yang\}@nyu.edu},}\\
  \small{\texttt{\{Daniel.Alber, eric.oermann\}@nyulangone.org}}\\
  \small{\texttt{\{alyakin314\}@gmail.com}}
}

\begin{document}
\maketitle
\begin{abstract}

\end{abstract}
Clinical measurements such as blood pressures and
respiration rates are critical in diagnosing and monitoring patient outcomes. It is an important component of biomedical data, which can be used to train transformer-based language models (LMs) for improving healthcare delivery. It is, however, unclear whether LMs can effectively interpret and use clinical measurements. We investigate two questions: First, can LMs effectively leverage clinical measurements to answer related medical questions? Second, how to enhance an LM's performance on medical question-answering (QA) tasks that involve measurements? We performed a case study on blood pressure readings (BPs), a vital sign routinely monitored by medical professionals. We evaluated the performance of four LMs: BERT, BioBERT, MedAlpaca, and GPT-3.5, on our newly developed dataset, BPQA (Blood Pressure Question Answering). BPQA contains $100$ medical QA pairs that were verified by medical students and designed to rely on BPs . We found that GPT-3.5 and MedAlpaca (larger and medium sized LMs) benefit more from the inclusion of BPs than BERT and BioBERT (small sized LMs). Further, augmenting measurements with labels improves the performance of BioBERT and Medalpaca (domain specific LMs), suggesting that retrieval may be useful for improving domain-specific LMs. \footnote{Our dataset is available through this link: \url{https://huggingface.co/datasets/Kekelilii/BPQA100}}

\begin{table*}[t]
\centering 
\begin{tabular}{p{4.5cm} p{11.5cm}}
\hline
\textbf{Question Type}                      & \textbf{Example}\\
\hline
Abnormality Detection under Special Context & \ldots an 80-year-old individual has a blood pressure reading of 155/65 mmHg. Is this considered hypertension?\\
\hline
Intervention Opinion  &  \ldots a patient with a blood pressure of 150/100 mmHg significantly reduces sodium intake, can this dietary change alone normalize their blood pressure?  \\
\hline
Symptoms and Illness  & \ldots a patient reports snoring heavily and feeling fatigued during the day. They have a blood pressure of 150/95 mmHg. Could sleep apnea be the cause?\\
\hline
Medical Research   & \ldots an individual with cerebrovascular disease, had an initial blood pressure of 160/100 mmHg and a high amino-terminal-pro-B-type natriuretic peptide (NT-proBNP) level. Will perindopril-based blood pressure-lowering therapy help with this individual’s situation? \\
\hline
\end{tabular}
\caption{Examples of four types of questions in BPQA.}
\label{tab:type}
\end{table*}

\section{Introduction}

Clinical measurements, such as blood pressure and respiration rate, are crucial in healthcare for accurate diagnosis and disease monitoring. Misinterpreting these measurements can be life-threatening. For example, blood pressure readings (BPs), which indicate the force of blood against the walls of arteries, is a vital sign that is measured for all patients. Abnormal BPs are associated with various diseases, such as cardiovascular disorders and kidney disease. Accurate interpretation of such measurements is crucial for patient health.  

Clinical measurements are ubiquitous in biomedical datasets which can be used to train and evaluate transformer-based language models (LMs). Many LMs \citep{singhal2022large, saab2024capabilities, yang2022gatortron, krishna-etal-2021-generating, nyutron} were trained and evaluated \citep{luo2022biored,jin2019pubmedqa,pampari2018emrqa, jin2020disease} on such datasets for healthcare applications. Appendix~\ref{apd:intro} shows that these data are rich in clinical measurements. 

It is unclear, however, whether LMs can effectively interpret and use clinical measurements. Although existing medical benchmarks evaluate LMs' capability on answering medical questions, they contain substantial additional information. For example, Appendix ~\ref{apd:intro} shows that MedQA, PubMedQA, emrQA contain both clinical measurements and additional contexts such as patient medical history, symptoms, and primary diagnoses (examples shown in Appendix~\ref{apd:adinfo}). This makes it challenging to isolate and assess LMs' performance on using clinical measurements alone, as LMs may rely on the other available information to answer questions. Previous studies on numerical reasoning over text have investigated the capability of LMs to understand and work with numbers \citep{dua-etal-2019-drop, wallace-etal-2019-nlp, thawani-etal-2021-representing, wu-etal-2021-math} and all kinds of measurements \citealp{park2022language}. However, none focus specifically on clinical measurements, which directly relate to health outcomes and requires domain knowledge to interpret. Focusing on clinical measurements allows a more targeted evaluation of LMs’ ability to use numerical data in a medical context.

Here we assessed how LMs (GPT-3.5 \citep{brown2020language}, MedAlpaca \citep{han2023medalpaca}, BERT \citep{devlin2018bert}, and BioBERT \citep{lee2020biobert}) can interpret and use BPs in medical question-answering (QA) tasks. We focused on BPs, a routinely monitored vital sign, as a case study because it is the most common clinical measurements. We selected the four models to compare the effect of different size, type (encoder or decoder), and pretrain corpus. We designed a new dataset called BPQA (Blood Pressure Question Answering) with QA pairs that are verified by medical students and designed to rely on BP. Using BPQA, we investigate the impact of BPs and their text label (low, normal, high) on the performance of the selected LMs in medical QA tasks. The results indicate: 1) GPT-3.5 (large sized LM) and MedAlpaca (medium sized LM) benefit more from the inclusion of BP measurements than BERT and BioBERT (small sized LMs), 2) augmenting labels to BPs improves performance for BioBERT and MedAlpaca (domain specific LMs), suggesting that retrieval may be useful for improving domain-specific LMs.

\section{Method}
We investigated the effect of adding BPs and labels (low, normal, high) on LMs' performances using BPQA, our newly designed medical QA dataset whose answers specifically depend on BPs. We compared the performances of four LMs: GPT-3.5, MedAlpaca, BERT, and BioBERT. Their sizes, types (encoder or decoder), pretrain corpus, and versions can be found in Appendix ~\ref{apd:ms}. The models were evaluated across four variants using accuracy.

\begin{table*}
  \centering
  \begin{tabular}{ll}
    \hline
    \textbf{Dataset}           & \textbf{Example} \\
    \hline
    BPQA      & \ldots blood pressure is \textbf{130/90} mmHg\ldots Answer Yes or No.\\
    BPQA-free     & \ldots blood pressure is \textcolor{red}{s/d} mmHg \ldots Answer Yes or No.\\
    BPQA-label       & \ldots blood pressure is 130/90 mmHg \textcolor{red}{(high)}\ldots Answer Yes or No.\\
    BPQA-free-label & \ldots blood pressure is \textcolor{red}{s/d} mmHg  \textcolor{red}{(high)} \ldots Answer Yes or No.\\
    BPQA-human-label & \ldots blood pressure for a pregnant woman is 130/90 mmHg  \textcolor{red}{(normal)} \ldots Answer Yes or No.\\
    \hline
  \end{tabular}
  \caption{Examples of BPQA dataset and its variants we created to assess the effect of adding BPs and augmenting labels. s/d is short for systolic/diastolic, whose definitions are in Appendix~\ref{apd:bpintro}.}
  \label{tab:variants}
\end{table*}

\begin{figure*}
\begin{subfigure}[t]{.45\textwidth}
  \includegraphics[width=\columnwidth]{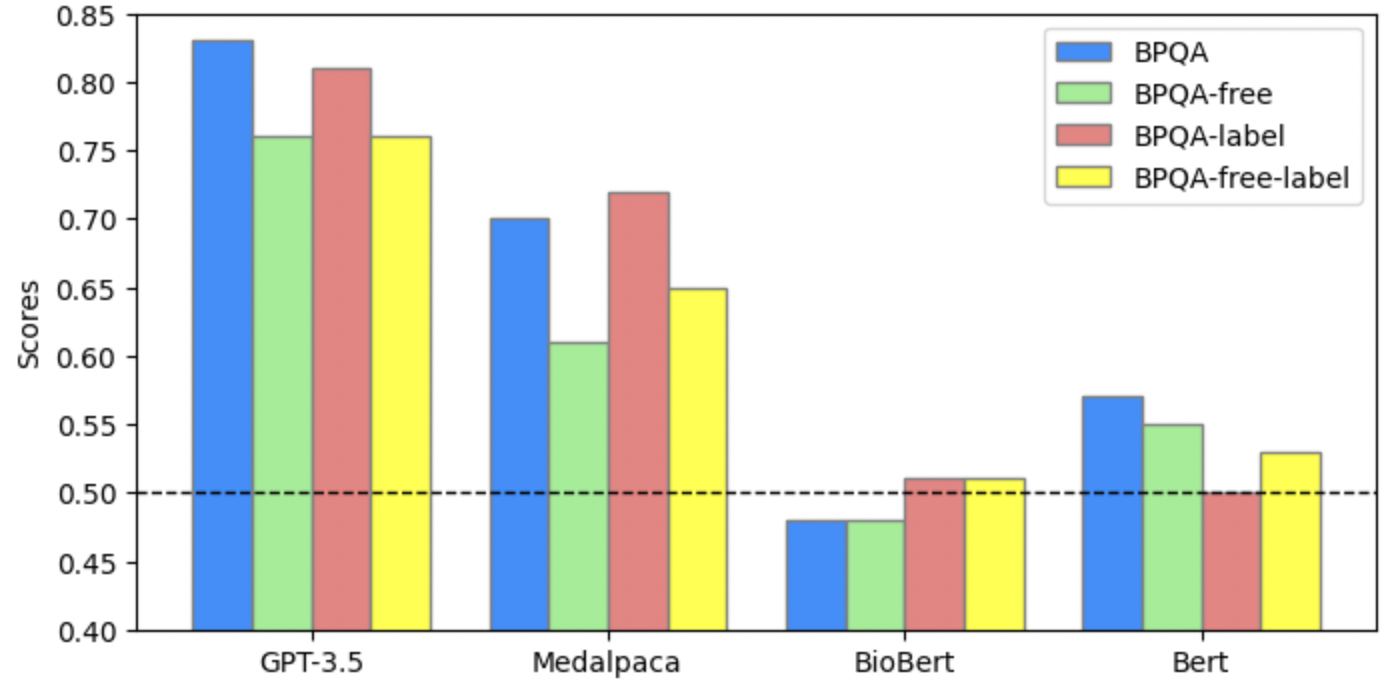}
  \caption{Zero-shot performance on BPQA shows that larger LMs benefit more from seeing BPs, label augmentation helps domain specific LMs, and GPT3.5 performs better with raw BPs.}
  \label{fig:result}
  \end{subfigure}
  \hfill
  \begin{subfigure} [t]{.45\textwidth}
  \includegraphics[width=\columnwidth]{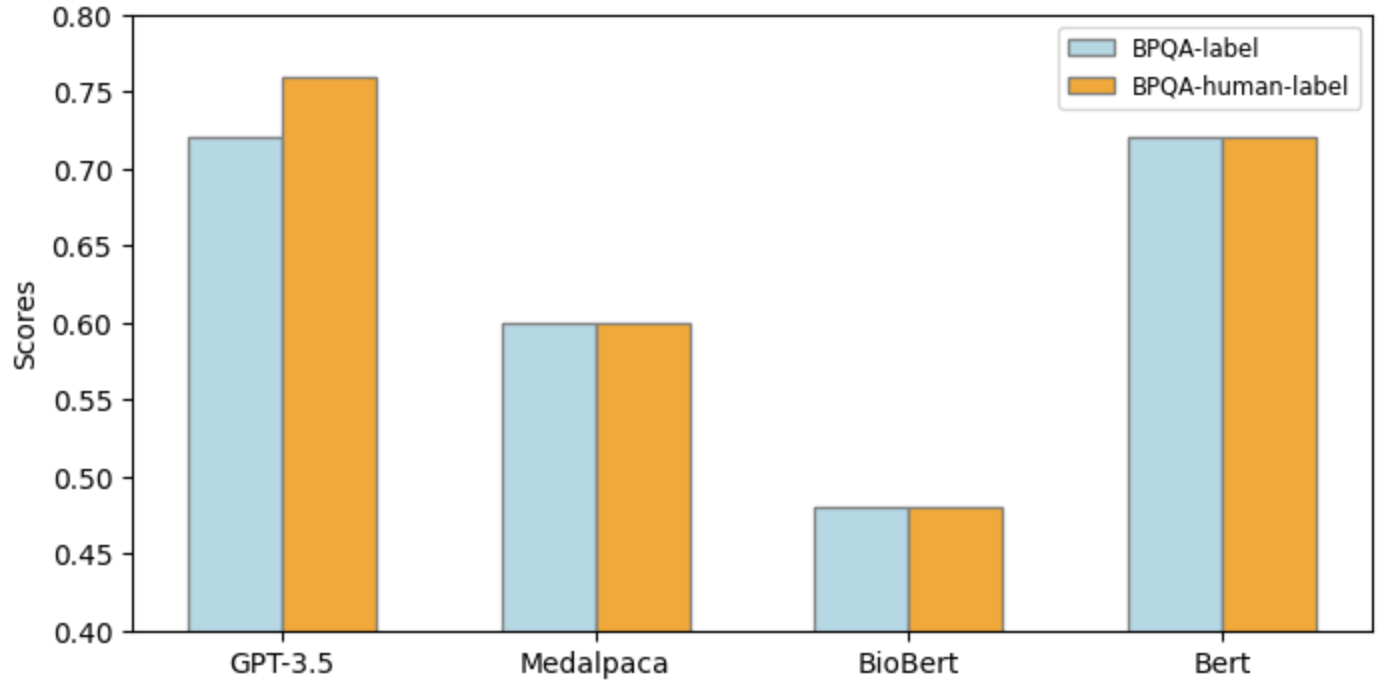}
  \caption{LMs Performance on the Special Context questions shows that GPT-3.5 benefits from context specific labels.}
  \label{fig:sc}
\end{subfigure}
\caption{Comparison of model performance on different BPQA variants.}
\end{figure*}

\subsection{Data}
We created a new dataset, BPQA which contains $100$ medical QA pairs verified by medical students and designed to rely on BPs.\footnote{Two senior medical students reviewed the dataset in May 2024.} All questions are formatted to be binary (answer ``yes'' or ``no''). The dataset is balanced, with $50$ questions having `yes'' answers and $50$ questions having ``no'' answers.

The dataset contains four categories of questions (25 QA pairs in each category) to evaluate LMs' ability of answering BPs-related questions in both clinical and research settings. The first three categories (Abnormality Detection under Special Context, Symptoms and Illness, and Intervention Opinion) are designed to assess LMs' ability to use BPs in clinical environments, similar to how doctors use BPs in patient care. These categories focus on identifying normal and abnormal BP ranges for specific populations, recognizing possible symptoms associated with specific BPs, and providing appropriate recommendations based on BPs, respectively. The fourth category, Medical Research, assesses LMs' ability to use BPs in research contexts, with manually selected and adapted BP-related questions from PubMedQA \citep{jin2019pubmedqa}. Table~\ref{tab:type} provides examples of the four question types.

We also created four variants of BPQA to assess the effect of adding BPs and adding BP labels (Examples shown in Table~\ref{tab:variants}). 
\begin{enumerate}\itemsep=-2pt 
 \item BPQA-free:  Replaced the BPs with s/d (systolic and diastolic).
 \item BPQA-label: Augmented categorical labels (low, normal, high) according to Centers for Disease Control and Prevention (CDC)\footnote{https://www.cdc.gov/high-blood-pressure/about/index.html}. 
 Specific BP threshold is in Appendix~\ref{apd:bpthreshold}. 
 \item BPQA-free-label: Replaced the BPs with s/d and augmented categofical label.
 \item BPQA-human-label: Augmented context specific categorical labels from our review. 
\end{enumerate}
In BPQA-human-label, the human assigned labels are used only for the Special Context questions. This category requires human labels because minority groups, such as pregnant women and infants, have BP evaluation different from CDC guidelines, while questions in other categories can be adequately labeled using CDC guidelines. For example, a blood pressure of 130/85 mmHg would be labeled as ``high'' in BPQA-label (according to CDC guidelines). However, this reading is considered ``normal'' for a pregnant woman. In the BPQA-human-label dataset, such a reading would be labeled as "normal" to account for the specific context of pregnancy. $13$ out of $25$ Abnormality Detection under Special Context questions was changed compare to BPQA-label.

\subsection{Experiments}
We performed zero-shot QA evaluation for selected LMs on all variants of BPQA. We chose zero-shot evaluation to measure LMs' inherent capabilities in using BPs, avoiding potential biases introduced by fine-tuning or few-shot learning. For BERT-based LMs, we used fill-mask classification with single token (details in Appendix~\ref{apd:fillmask}); for MedAlpaca and GPT-3.5, we used text-generation with [Yes/No] choices in the prompts. 

\section{Results}

\textbf{GPT-3.5 and MedAlpaca (a large and a medium sized LM) benefit more than BERT and BioBERT (small sized LM) from seeing BPs}. Figure~\ref{fig:result} shows that GPT-3.5, MedAlpaca, and BERT achieve higher accuracy (9\%, 15\%, 3\% gain respectively) when tested on the BPQA (blue bar) compared to the BPQA-free (green bar). However, BioBERT (third column) shows no change between the blue and green bars. This suggests that larger LMs may leverage BPs more effectively when answering medical questions, while smaller LMs may benefit less or do not utilize BPs at all.

Although all the questions are designed to rely solely on BPs, \textbf{some LMs achieve above random accuracy for BPs-redacted questions}, possibly because they used contextual information in the question as a spurious correlation. The green bars in Figure~\ref{fig:result} show that BERT, MedAlpaca, and GPT-3.5 all reach accuracy above 0.5 (dashed line) on BPQA-free (green bar, BPs removed). We speculate that BP-related tokens (such as ``old'' and ``pregnant'' because they are risk factors for hypertension ) in the questions may contribute to this. 

LMs' performances vary on the BPQA-label dataset: \textbf{Seeing BPs with label augmentation helps BioBERT and MedAlpaca (domain-specific LMs), but does not help or hurts BERT and GPT-3.5 (small and large sized general LMs)}. Figure~\ref{fig:result} shows that adding BP labels (red bar) improves BioBERT and MedAlpaca's performances from original (blue bar) by 6\% and 3\% respectively.
However, label augmentation hurts BERT's performance by 12\% (the red bar is lower than the blue bar). We speculate this drop might be related to the increased sentence complexity from the inserted label. GPT-3.5's performance remains largely unchanged with a 1\% decrease from label augmentation (blue versus red bar). 

\textbf{GPT-3.5 peforms better with raw BPs and without CDC labels}. Focusing on the first column of Figure~\ref{fig:result}, we compare GPT-3.5's performance on datasets with (blue and red bars) and without (green and yellow bars) raw BPs. The best performance (0.83) comes from original dataset with raw BPs (blue bar). The second best (red bar) has a 1\% drop with augmented CDC label. And the worst peformance (0.76) are variants without BPs (green and yellow bar). This suggests GPT-3.5's performance hurt from including CDC labels and benefits from seeing raw measurements.

\textbf{Context-specific labels helps GPT-3.5 achieve higher accuracy.} Focusing on the Special Context questions (questions related to minority groups like pregnant women, whose BP evaluation might be different from CDC guidelines) in Figure~\ref{fig:sc}, GPT-3.5 improves its performance by 6\% on BPQA-human-label compared to BPQA-label (orange versus blue bar) after human-labeling to conform to the specific contexts. This shows the incompatibility between BPs under special contexts and the general CDC labels might have hurt GPT3.5's performance on BPQA-label. For example, 130/85 mmHg is labeled ``high'' by CDC but is actually ``normal'' for a pregnant woman. With context-specific labeling, GPT-3.5 benefits from the label augmentation. Please refer to Appendix~\ref{apd:human} for detailed explanation on BPQA-human-label result.

\section{Discussion} 

The positive effect of context-specific labels suggests that patient-context-aware \textbf{retrieval augmentation may improve clinical language model's generalization to minority patients}, such as pregnant women with unique needs and vitals. While we manually modified CDC labels to special contexts, retrieval models \citep{Zakka2024-ak} can incorporate relevant contextual information from specialized database during the training and inference, helping LMs better understand and respond to the variability of different demographics. The negative effect of using general CDC labels also indicates the importance of accounting for the specific needs of different patient demographics.

Our work highlights the \textbf{need for more specialized benchmarks for clinical LMs}. In our case, the skill of understanding clinical measurements is crucial for practical use, but our early experiments showed that existing benchmarks such as MedQA involve too much additional information to target clinical measurements (see Appendix~\ref{apd:medqa} and Appendix~\ref{apd:adinfo} for details). Our BP-targeted BPQA dataset showed the disparity in LMs' abilities to leverage BPs and the room for improvements. While BPQA serves as an initial step, we need more specialized benchmarks for targeted evaluation of LMs' skills in interpreting and reasoning over quantitative medical data. 

\textbf{A future direction for improving LM's skills with clinical measurements is modifying tokenizers}. The unique numeric format in medical contexts, like special characters and units (e.g. 130/85 mmHg), may benefit from specialized tokenizers. Most existing tokenizers \citep{kudo2018sentencepiece, sennrich2016neural} are designed for natural text and may struggle with such formatted numerical representations while preserving quantitative semantics. 

\section{Limitations}
Our study used BPs as a representative for clinical measurement, which does not fully capture the diversity of clinical scenarios. We also did a very preliminary study with a synthetic dataset (only $100$ QA pairs) on $4$ LMs. Future works include improving the dataset's breadth by including more types of measurements, more rigorously testing our finding with more LMs, testing the effect of retrieval model on tasks involving minority groups, and designing tokenizers that work well with clinical measurements.


\newpage
\bibliography{custom}

\clearpage
\appendix

\section{Synthetic BPQA datasets}\label{apd:bp}
\subsection{BP Introduction}\label{apd:bpintro}
BP is recorded as two numbers, in the form of systolic/diastolic mm Hg (millimeters of mercury): Systolic blood pressure indicates how much pressure blood is exerting against your artery walls when the heart contracts. Diastolic blood pressure indicates how much pressure blood is exerting against your artery walls while the heart muscle is resting between contractions.

\subsection{BP Threshold}\label{apd:bpthreshold}
The threshold we used for labeling BP refers to Centers for Disease Control and Prevention (CDC). 

\begin{table}[hbtp]
  \begin{tabular}{lll}
  \hline
  \bfseries Systolic BP (s) & \bfseries Diastolic BP (d) & \bfseries Label\\
  \bfseries (mmHg) & \bfseries (mmHg) &  \\
  \hline
  s $<$ 90 &  & Low\\
  \hline
    & d $<$ 60 & Low\\
  90  $\leq$ s $<$ 120  & 60  $\leq$ d $<$ 80  & Normal\\
    & d $\geq$ 80 & High\\
  \hline
  s $\geq$ 120 &   & High\\
  \hline
  \end{tabular}
  {\caption{BP Threshold}}
\end{table}

\section{Examples of Datasets Involving Clinical Measurements}\label{apd:intro}
Table~\ref{tab:intro} shows that biomedical data used to train and evaluate LMs are rich in clinical measurements. 
\begin{table*}
\centering 
\begin{tabular}{p{5cm} p{11cm}}
\hline
\textbf{Dataset}           & \textbf{Example}\\
\hline
PubMed abstracts & \ldots estimation error of \textbf{-2.06 ± 6.89 mmHg} for systolic BP, and \textbf{0.89} and  \textbf{-4.66 ± 4.91 mmHg} for diastolic BP\ldots  \\
\hline
MIMIC-III \citep{johnson2016mimic} & includes vital signs, medications, \textbf{laboratory measurements}, and more.\\
\hline
MedQA  & \ldots vital signs are: blood pressure, \textbf{148/90 mm Hg}, heart rate, \textbf{88/min} \ldots \\
\hline
PubMedQA  & \ldots mortality rates doubled at < \textbf{100 mm Hg}, tripled at < \textbf{90 mm Hg} and were 5- to 6-fold at < \textbf{70 mm Hg}, irrespective of age \ldots \\
\hline
emrQA & \ldots physical examination: BMI:\textbf{33.4}. Pulse: \textbf{60}. resp. rate: \textbf{18}\ldots \\
\hline
\end{tabular}
\caption{Examples of datasets involving clinical measurements}
\label{tab:intro}
\end{table*}

\section{Substantial Information Contained in Example Questions of MedQA, PubMedQA, and emrQA}\label{apd:adinfo}

\begin{enumerate}
 \item MedQA:  A 27-year-old male presents to urgent care complaining of \textcolor{red}{pain with urination}. He reports that the pain started 3 days ago. \textcolor{red}{He has never experienced these symptoms before. He denies gross hematuria or pelvic pain.} He is sexually active with his girlfriend, and they consistently use condoms. \ldots His mother has rheumatoid arthritis. \textbf{The patients temperature is 99 F (37.2 C), blood pressure is 112/74 mmHg, and pulse is 81/min.} On physical examination, there are no lesions of the penis or other body rashes. No costovertebral tenderness is appreciated. \textcolor{red}{A urinalysis reveals no blood, glucose, ketones, or proteins but is positive for leukocyte esterase. A urine microscopic evaluation shows a moderate number of white blood cells but no casts or crystals. A urine culture is negative.} Which of the following is the most likely cause for the patient’s symptoms? A: Chlamydia trachomatis, B: Systemic lupus erythematosus, C: Mycobacterium tuberculosis, D: Treponema pallidum

 \item PubMedQA:  \ldots Associations were assessed by logistic regression with respect to systolic, diastolic and pulse pressure, with adjustment for education, work status, physical activity, smoking, body mass and lipid levels. \ldots In the prospective study of disease-free women, \textcolor{red}{baseline pulse pressure and systolic pressure were inversely associated with risk of low back pain} \textbf{[odds ratio (OR) 0.93 per 10 mmHg increase in pulse pressure, 95\% confidence interval (CI) 0.89-0.98, p=0.007; OR 0.95 per 10 mm Hg increase in systolic pressure, 95\% CI 0.92-0.99, p==0.005.]} \ldots Does high blood pressure reduce the risk of chronic low back pain?
 \item emrQA: 08/31/96 ascending aortic root replacement with homograft with omentopexy. The patient continued to be hemodynamically stable making good progress. Physical examination: \textbf{BMI: 33.4} \textcolor{red}{Obese, high risk}. \textbf{Pulse: 60. resp. rate: 18}. Has the patient ever had an abnormal BMI?
\end{enumerate}

Here are example questions from MedQA, PubMedQA, and emrQA. While these questions contain clinical measurements (highlighted in bold), they often provide additional context (shown in red) that allow LMs to infer the answer without directly interpreting or using the measurements.

\newpage
\section{Model Selection}\label{apd:ms}
Our model selection allows us to observe the performance of models with differing size, type, and pre-training data. See table ~\ref{tab:model} for sizes and type of selcted models.
BERT is trained on Wikipedia articles and Book Corpus. BioBERT, on top on Bert, is trained on biomedical articles from PubMed abstracts. MedAlpaca is trained on a variety of medical texts, encompassing resources such as medical flashcards, wikis, and dialogue datasets. GPT-3.5 is trained on a large and diverse of data like webtexts, books, and Wikipedia. Here, we used \textbf{gpt-3.5-turbo-instruct} whose training data was updated up to Sep 2021.
\begin{table}
\begin{tabular}{lll}
\hline
\textbf{Model} & \textbf{Size} &\textbf{type}\\
\hline
Bert & 110M & encoder\\
BioBERT  & 110M & encoder\\
MedAlpaca & 6.74B & decoder\\
GPT-3.5-turbo-instruct  & NA & decoder\\
\hline
\end{tabular}
\caption{We selected encoders and decoders models of different sizes and pretrain corpus}
\label{tab:model}
\end{table}

\section{Fill-mask Classification for BERT Based Model}\label{apd:fillmask}
For BERT and BioBERT, we used fill-mask classification with single token. The BPQA datasets in ~\ref{tab:variants} are modified by removing ``Answer Yes or No'' and adding ``The answer is [MASK]''. The candidates for [MASK] are constrained to ``yes'' and ``no''. 

\newpage
\section{Full Evaluation Results}\label{apd:fullresults}
\begin{table}[hbtp]
\begin{tabular}{lllll}
\hline
\bfseries Model & \bfseries BPQA & \bfseries BPQA & \bfseries BPQA & \bfseries BPQA\\
& & \bfseries -free & \bfseries -label &  \bfseries -free-label\\
\hline
GPT-3.5 & 0.83 & 0.76 & 0.82 & 0.76\\
MedAlpaca & 0.70 & 0.61 & 0.72 & 0.65\\
BioBERT & 0.48 & 0.48 & 0.51 & 0.51\\
Bert & 0.57 & 0.55 & 0.50 & 0.53\\
\hline
\end{tabular}
\caption{Evaluation results of different LMs on various BPQA datasets.}
\label{tab:evaluation}
\end{table}

\section{BPQA-human-label Result Explanation}\label{apd:human}
13 out of 25 questions in the Special Context group were modified based on the need to align with specific question contexts. Focusing on the 13 modified questions, the performances of BERT, BioBERT and MedAlpaca showed no change before and after the modification. GPT-3.5 increased its performance from 0.61 to 0.69, indicating that GPT-3.5 is able to detect the trivial change and benefit from context-specific labeling.

\newpage
\section{Experiments using MedQA-USMLE}\label{apd:medqa}
On the early stage of our study, we extracted 3500 questions that contain BPs in the MedQA-USMLE benchmark as the evaluation dataset (\textbf{MedQA-BP}), and created a variant with BPs removed (\textbf{MedQA-BP-free}). We tested BERT, BioBERT and GPT-3.5 on the two datasets. Looking at the results in Table~\ref{tab:medqa}, the largest performance gap for all the models between MedQA-BP and MedQA-BP-Free is 0.01, which we believe is not significant enough to support any claims. This might be because questions in MedQA contain abundant additional information besides BPs, so adding or removing single BPs can only cause trivial differences on model performances. In this case, our experiments cannot be optimally employed on MedQA. Thus, we developed a synthetic BP-targeted dataset to display more significant results. 

\begin{table}[hbtp]
\begin{tabular}{lll}
\hline
\bfseries Model &  \bfseries MedQA-BP & \bfseries MedQA-BP-Free\\
\hline
GPT-3.5 & 0.521 & 0.531\\
BioBERT & 0.262 & 0.252\\
Bert & 0.242 & 0.249\\
\hline
\end{tabular}
\caption{Result on MedQA-BP}
\label{tab:medqa}
\end{table}

\end{document}